\documentclass{article}

\usepackage{arxiv}

\usepackage[utf8]{inputenc} 
\usepackage[T1]{fontenc}    
\usepackage{hyperref}       
\usepackage{url}            
\usepackage{booktabs}       
\usepackage{amsfonts}       
\usepackage{nicefrac}       
\usepackage{microtype}      
\usepackage{lipsum}
\usepackage{graphicx}
\graphicspath{ {./images/} }

\usepackage[table]{xcolor}
\usepackage{graphicx}
\usepackage{amssymb,amsmath}
\usepackage[utf8]{luainputenc}
\usepackage[]{algorithm2e}
\usepackage{arydshln}
\usepackage{floatrow}
\usepackage{ctable}
\usepackage{multirow}
\usepackage{hhline}
\usepackage{dsfont}
\usepackage{pifont}
\usepackage[section]{placeins}

\newfloatcommand{capbtabbox}{table}[][\FBwidth]

\title{S-MULTI-SNE: Semi-Supervised Classification and Visualisation of Multi-View Data}

\author{
 Theodoulos Rodosthenous\thanks{Contact author} \\
  Department of Mathematics\\
  Imperial College London\\
  London, SW7 2AZ, UK \\
  \texttt{tr1915@ic.ac.uk} \\
   \And
 Vahid Shahrezaei \\
  Department of Mathematics\\
  Imperial College London\\
  London, SW7 2AZ, UK \\
  \texttt{v.shahrezaei@imperial.ac.uk} \\
  \And
 Marina Evangelou \\
  Department of Mathematics\\
  Imperial College London\\
  London, SW7 2AZ, UK \\
  \texttt{m.evangelou@imperial.ac.uk} \\
}

\begin{document}
\maketitle
\begin{abstract}
An increasing number of multi-view data are being published by studies in several fields. This type of data corresponds to multiple data-views, each representing a different aspect of the same set of samples. We have recently proposed multi-SNE, an extension of t-SNE, that produces a single visualisation of multi-view data. The multi-SNE approach provides low-dimensional embeddings of the samples, produced by being updated iteratively through the different data-views. Here, we further extend multi-SNE to a semi-supervised approach, that classifies unlabelled samples by regarding the labelling information as an extra data-view. We look deeper into the performance, limitations and strengths of multi-SNE and its extension, S-multi-SNE, by applying the two methods on various multi-view datasets with different challenges. We show that by including the labelling information, the projection of the samples improves drastically and it is accompanied by a strong classification performance. 
\end{abstract}


\section{Introduction} \label{introduction}

Multi-view data are usually described as a collection of data taken from different sources on the same samples. It is now very common  for multi-view data to be generated in different fields; for example multi-omics datasets in biomedical studies \cite{rodosthenous2020}, crystal structure data in the field of chemistry \cite{cheng2020}, data sources in social science \cite{holmbeck2002} and cyber-security \cite{kent2015}. In biomedical studies multiple omics datasets, {\em e.g. proteomics, genomics, transcriptomics}, are generated on the same individuals. Through these studies the researchers are interested in understanding the relationships between the omics datasets, the underlying biology, and also enhance their relationship with the studied disease ({\em e.g.} classifying patients as healthy or not). In this manuscript, we focus on the latter, and specifically on the task of classifying samples by utilising the multi-view data. We propose a semi-supervised learning approach, named {\em S-multi-SNE}, that incorporates the labelling information of the training set alongside the multi-view data (training and test) to visualise all samples and classify the labels of the test set. 


S-multi-SNE is an extension of our recent work on multi-view visualisation approach, {\em multi-SNE} that produces a single representation of the samples by incorporating the information of all data views. {\em Multi-SNE} is a multi-view extension of the widely used dimensionality reduction approach, t-distributed Stochastic Neighbour Embedding (t-SNE) \cite{tsne}, that has gained great popularity over the last years as it provides a comprehensible low-dimensional projection of the samples in a single-view setting. Multi-SNE was found to have superior performance in the visualisation of the samples and identification of any underlying structure when compared with the competitive extension of t-SNE, named {\em m-SNE} \cite{Xie2011_mSNE}, and other multi-view manifold learning approaches \cite{rodosthenous2021}.

The proposed adaptation of the multi-SNE approach focuses more in producing a good data visualisation by incorporating the labelling information of the samples, which is constructed as a binary matrix of size $N \times C$, where $C$ presents the total number of classes of the samples and $N$ is the number of samples. The cell $(i,j)$ of the matrix takes the value $1$ if sample $i$ belongs to class $j$ and zero otherwise. For example, in a multi-omics experiment on cancer patients and controls, the labelling matrix will have two columns, one column for the cancer patients, and another one for the controls. Similarly, in a cancer subtypes study where the samples are patients with different cancer types (for example \cite{Wang2014}), each column of the labelling matrix will correspond to each cancer type.

The proposed approach, {\em S-multi-SNE}, combines labelling information of the training samples, with the training and test sets of the multi-view data, for classifying the labels of the test samples. This is done by applying the multi-SNE algorithm on the available data and applying a classification algorithm on the projected low-dimensional embeddings produced by the algorithm. {\em S-multi-SNE} is a transductive algorithm, as it does not build a generic predictive model. Such algorithms tend to make predictions on a specific test set \cite{stanescu2016}. If a new data point is added to the test set, then the algorithm has to re-run from the beginning to train the model and then to predict the labels. Transductive learning algorithms are preferred when multiple test (query) sets are available with different characteristics. 

The {\em S-multi-SNE} projections are treated as features in the classification algorithm that predicts the classes of the test samples. Different classification algorithms can be utilised for this purpose. Through a series of experiments we illustrate that the K-Nearest Neighbours (KNN)\cite{knn1951} classification algorithm has a good performance. An advantage of KNN is that a good classification score ensures a good visualisation of the data. That is, because KNN separates different classes into neighbourhoods and classifies the samples in the test set by looking at their neighbours \cite{knn1951}.

\subsection*{Related work} \label{relatedWorks}

In the last few years, a number of multi-view semi-supervised learning approaches have been proposed \cite{nie2016, nie2017, bo2019, yang2021}. Bo et al. (2019) \cite{bo2019} conducted a simulation study where they compared the performance of the recently published semi-supervised multi-view classification approaches: Auto-weighted Multiple Graph Learning (AMGL)\cite{nie2016}, Multi-view Learning with Adaptive Neighbors (MLAN) \cite{nie2017} and Latent Multi-view Semi-Supervised Classification (LMSSC)\cite{bo2019}. In their study, Bo et al.( 2019) demonstrated that LMSSC was superior to the other algorithms, under different scenarios. The LMSSC approach classifies the test samples in two steps: (a) constructs a graph, with samples as nodes and weighted edges based on similarities among all data-views, and (b) uses label propagation to infer the labels on unlabelled samples. Following the results of Bo et al. (2019), we have compared our proposed {\em S-multi-SNE} approach with the LMSSC approach. 

The popularity of t-SNE attracted many researchers who proposed several variations and extensions of the algorithm. Recently, Cheng et al. (2020) proposed St-SNE \cite{cheng2020_stSNE}, a supervised extension of t-SNE. Similarly to our proposal, St-SNE considers the labelling information as an additional data-view. In contrast to S-multi-SNE, the unlabelled samples are classified differently. They proposed three strategies for classification, one of which uses Neural Networks (nSt-SNE) and it is similar to the parametric t-SNE \cite{vanDerMaaten_parametric}. Another strategy (dSt-SNE) implements St-SNE twice to predict the classes of unlabelled samples. The latter strategy was used to compare a single-view approach (St-SNE) against a multi-view approach (S-multi-SNE) and to highlight the benefits of incorporating multiple data-views in the analysis. 

A comparative study between S-multi-SNE, LMSSC and St-SNE was conducted to explore and assess their classification performance. All three algorithms were implemented on 10\%, 20\%, 50\%, and 80\% of the samples in training, covering both semi-supervised and supervised scenarios. The focus of the comparison with St-SNE lies on 80\% training rate, while the emphasis against LMSSC falls on the low training rates (10\%, 20\%, 50\%). We show that S-multi-SNE performs closely to LMSSC and outperforms St-SNE. In particular, S-multi-SNE was superior on 80\% training/test rate and it was found to outperform LMSSC on datasets with imbalanced labels or small sample size per class. 	

In the following section, we describe the proposed {\em S-multi-SNE} approach and the multi-view datasets used. Through a series of experiments we illustrate and discuss:

\begin{itemize}
	\item the performance of visualising samples when labelling information is included
	\item the choice of the classifier algorithm and how it affects the performance of S-multi-SNE
	\item the performance of S-multi-SNE versus LMSSC and St-SNE
	\item the performance of S-multi-SNE and LMSSC on datasets with imbalanced labels and small sample size per class
\end{itemize}

\section{Materials and Methods} \label{supervised_multi-SNE}

In this section, multi-SNE is described and its extension, S-multi-SNE is introduced. The classification algorithms and datasets used in this study are then reported.

\subsection{Multi-SNE}	\label{multi-SNE}

Suppose that $\textbf{X}$ is a multi-view dataset that contains $M$ data-views. Let $X^{(m)}\in \mathbb{R}^{N \times p_m}$ denote the $m^{th}$ data-view, with $\textbf{x}_i^{(m)}$ being the $i^{th}$ data point of $X^{(m)}$, where $m = 1, \cdots, M$. Let $Y \in \mathbb{R}^{N \times d}$ represent the low-dimensional embedding of the original data obtained as the output of multi-SNE; $\textbf{y}_i$ is the $i^{th}$ data point of $Y$ and $d=2$ was set throughout the paper.

For data-view $m$, multi-SNE measures the probability distribution, $P^{(m)}$, of each data point, $\textbf{x}_i^{(m)}$, as follows: For every sample $i$, a sample $j$ is taken as its potential neighbour with probability $p_{ij}^{(m)}$, given by:

\begin{align}
	p_{ij}^{(m)}	= \frac{\exp{(-(d^{(m)}_{ij})^{2})}}{ \sum_{k \neq i} \exp{(-(d^{(m)}_{ik})^{2})} } ,
\end{align}
where $d^{(m)}_{ij} = \frac{||\textbf{x}^{(m)}_i - \textbf{x}^{(m)}_j||^2}{2 \sigma_i^2}$ represents the dissimilarity between points $\textbf{x}^{(m)}_i$ and $\textbf{x}^{(m)}_j$. The obtained probability distribution $P^{(m)}_i = \sum_j p^{(m)}_{ij}$ has a fixed perplexity, which refers to the effective number of local neighbours. Perplexity is defined as $Perp(P^{(m)}_i)=2^{H(P^{(m)}_i)}$, where $H(P^{(m)}_i)=-\sum_j p^{(m)}_{ij} \log_2 p^{(m)}_{ij}$ is the Shannon entropy of $P^{(m)}_i$, typically taking values between $5$ and $50$. The results presented in this manuscript are taken with optimized perplexity.

A probability distribution in the low-dimensional space follows Student's t-distribution with one degree of freedom \cite{tsne} and it is computed as follows:
\begin{align*}
	q_{ij}	= \frac{ (1 + ||\textbf{y}_i - \textbf{y}_j||^2)^{-1} }{ \sum_{k \neq l} (1 + ||\textbf{y}_k - \textbf{y}_l||^2)^{-1} } , 
\end{align*}
which represents the probability of point $i$ selecting point $j$ as its neighbour. 

The Kullback-Leibler divergence (KL-divergence) provides a measure of how different a probability distribution, $G$ is from a second probability distribution, $H$, denoted by $KL(G||H)$ \cite{kullback1951}. If $KL(G||H) = 0$, then the probability distributions $G$ and $H$ are identical. The induced embedding output, $\textbf{y}_i$, represented by probability distribution, $Q$, is obtained by minimising the sum of all KL-divergence measures between $Q$ and the distributions of every data-view $m = 1, \cdots, M$.  In other words, multi-SNE minimises the cost function given by equation (\ref{eq:multiSNE_cost}).

\begin{align}
	C_{multi-SNE} = & \sum_m \sum_i w^{(m)} KL(P^{(m)}_i||Q_i) \nonumber\\
	= & \sum_m \sum_{i} \sum_{j}  w^{(m)} p^{(m)}_{ij} \log \frac{p^{(m)}_{ij}}{q_{ij}} \label{eq:multiSNE_cost} , 
\end{align}
where $w^{(m)}$ provides a weight value for each data-view, and $\sum_m w^{(m)}=1$. Here, we have taken equal weights on all data-views, \textit{i.e.} $w^{(m)} = \frac{1}{M}, \quad \forall m=1, \cdots M$.

\subsection{S-multi-SNE} \label{S-multi-SNE}


The iterative property of multi-SNE provides the option to modify the algorithm in a way that includes the labelling information and consequently make predictions on unlabelled samples. In this section, we propose such a modification, named \textit{S-multi-SNE}. The algorithm of this approach can be found in the supplementary material.

Suppose that $M$ data-views are available, and assume w.l.o.g. that the last data-view, denoted by $X^{(l)} = X^{(M)}$, contains the labelling information in a binary matrix format. For $m = 1, \cdots, M$, let $X^{(m)}_{TR} \subset X^{(m)}$ be the training set (contains information on labelled samples) and $X^{(m)}_{TE} \subset X^{(m)}$ be the test set (with unlabelled samples).  The low-dimensional embeddings of the data points in the training set are computed by using all available data-views, including $X^{(l)}_{TR}$. On the other hand, the embeddings of the data points in the test set do not consider $X^{(l)}_{TE}$, since that information is missing ($X^{(l)}_{TE} = \emptyset$). 

Let $\mathfrak{I}^{(l)} \in \mathbb{R}^{N \times N}$ be defined by:
\begin{align}
	\mathfrak{I}^{(l)}_{ij}  = \begin{cases} 
		1 & \text{if } \{D^{(l)}_{i} = 1 \} \land \{D^{(l)}_{j} = 1 \} \\
		0 & \text{otherwise}
	\end{cases} \nonumber
\end{align}
where $D^{(l)} \in \mathbb{R}^{N}$ denotes the \textit{missing data}, defined by:
\begin{align}
	D^{(l)}_{i}  = \begin{cases} 
		0 & \text{if } \textbf{x}_i^{(l)} \text{ is missing} \\
		1 & \text{if } \textbf{x}_i^{(l)} \text{ is observed} 
	\end{cases} \nonumber
\end{align}
The cost function of S-multi-SNE is given by:
\begin{align}
	C_{S-multi-SNE} = & \left[  \sum_m^{M-1} \sum_{i} \sum_{j}  w^{(m)} p^{(m)}_{ij} \log \frac{p^{(m)}_{ij}}{q_{ij}} + \right. \nonumber \\
	+ & \left. \mathfrak{I}^{(l)}_{ij} w^{(l)} p^{(l)}_{ij} \log \frac{p^{(l)}_{ij}}{q_{ij}} \right]  
\end{align}

In every experiment of this study, the data were normalised via PCA, before the implementation of S-multi-SNE. In particular, for each data-view, the first $c$ principal components (PC) that describe 80\% of the variance were taken as input in the algorithm. This dimensionality reduction pre-processing step speeds up the algorithm and suppresses some noise, without distorting the distances between data points. Normalisation via PCA is commonly used as a pre-processing step for t-SNE, since it was implemented in the original proposal of the algorithm \cite{tsne}.

T-SNE and by extension its variations, including S-multi-SNE, require a perplexity value, which needs to be tuned. All projections presented in this manuscript are taken with optimised perplexity, which was selected qualitatively by reviewing the data visualisations of each method on a range of perplexity values, $S = \left\lbrace 2, 10, 20, 50, 80, 100, 200 \right\rbrace$. A quantitative evaluation on the classification task assisted in identifying the optimised perplexity. 




\subsection{Classification algorithms} \label{classifiers}

The output of S-multi-SNE, $Y \in \mathbb{R}^{N \times d}$, are low-dimensional embeddings of all samples, including the ones in the test set. These projections can then be treated as input features into classification algorithms. In a practical manner, any general-purpose classifier can be used to predict the classes of the unlabelled samples. In this paper, several standard classifiers were explored: (a) Support Vector Machine (SVM) \cite{svm1995}, (b) Linear Discriminant Analysis (LDA) \cite{lda1936}, (c) Decision Trees (DT) \cite{decisionTrees1986}, (d) Random Forests (RF) \cite{randomForests2001}, (e) Neural Network, via Multi-Layer Perceptron (NN) \cite{mlp1986}, and (f) K-Nearest Neighbours (KNN) \cite{knn1951}. 

Each of these classifiers require tuning of one or more parameters. For example, different kernel functions were explored in SVM, and and different solver functions were assessed in LDA. The number of trees, forests, layers and neighbours were optimised in DT, RF, NN and KNN, respectively.  A grid search cross-validation framework was implemented to tune the parameter values in each classifier. 

The following steps were taken to test the performance of the classifiers on S-multi-SNE, applied on a multi-view dataset $\textbf{X}$.

\begin{enumerate}
	\item Randomly split the samples in training/test sets with 10\%, 20\%, 50\% and 80\% of samples lying in the training set. The split is performed in proportion to each class size within a dataset. 
	\item Implement S-multi-SNE on \textbf{X}.
	\item Implement a classifier algorithm (\emph{e.g.} KNN) on the low-dimensional embeddings produced by S-multi-SNE to classify the samples in the test set. 
	\item Repeat steps $1-3$, for a $N_{iter} = 100$ times.
\end{enumerate}

\subsection{Data Description} \label{dataDescription}

The aim of the study is to explore the performance of S-multi-SNE and compare it against existing approaches. In order to test the robustness of the algorithm, it is important to explore datasets with distinct attributes. Four real and one synthetic datasets were analysed. The different characteristics (\emph{e.g.} high-dimensionality, heterogeneity, number of data-views, samples and classes) of each dataset  allow us to evaluate the methods in a range of real-life situations with noisy data. The real datasets are classified as heterogeneous due to the nature of their data; the synthetic dataset is classified as non-heterogeneous, since it was generated under the same conditions and distributions. The datasets analysed in this paper are described below:\\

\textbf{Handwritten Digits \footnote{\href{https://archive.ics.uci.edu/ml/datasets/Multiple+Features}{https://archive.ics.uci.edu/ml/datasets/Multiple+Features}} \cite{Dua:2019}:} Extracted from a collection of Dutch utility maps. \\ \emph{Number of classes}: $\textbf{10} \left[ \text{Handwritten numerals } (0-9) \right] $.\\ \emph{Number of data-views}: $\textbf{6}$: (a) Fourier coefficients of the character shapes ($p_1 = \mathit{76}$), (b) profile correlations ($p_2 = \mathit{216}$), (c) Karhunen-Love coefficients ($p_3 = \mathit{64}$), (d) pixel averages in 2 x 3 windows ($p_4 = \mathit{240}$), (e) Zernike moments ($p_5 = \mathit{47}$) and (f) morphological features ($p_6 = \mathit{6}$).\\ \emph{Number of samples}: $\textbf{2000} \left[ \textit{200} \text{ patterns per class }\right]$. 

\textbf{Caltech7 \footnote{\label{mvData}\href{https://github.com/yeqinglee/mvdata}{https://github.com/yeqinglee/mvdata}} \cite{caltech101}:} Subset of Caltech-101. \\ \emph{Number of classes}: $\textbf{7} \left[ \text{Pictures of 7 different objects}  \right] $.\\ \emph{Number of data-views}: $\textbf{6}$: (a) Gabor ($p_1 = \mathit{48}$), (b) wavelet moments ($p_2 = \mathit{40}$), (c) CENTRIST ($p_3 = \mathit{254}$), (d) histogram of oriented gradients ($p_4 = \mathit{1984}$), (e) GIST ($p_5 = \mathit{512}$), and (f) local binary patterns ($p_6 = \mathit{928}$).\\ \emph{Number of samples}: $\textbf{1474}$: Imbalanced dataset with samples per class: \{A: 435, B: 798, C: 52, D: 34, E: 35, F: 64, G: 56\}. 


\textbf{Cancer Types \footnote{\href{http://compbio.cs.toronto.edu/SNF/SNF/Software.html}{http://compbio.cs.toronto.edu/SNF/SNF/Software.html}} \cite{Wang2014}:} Multi-omics.\\ \emph{Number of classes}: $\textbf{3} \left[ \text{Cancer types (breast, kidney, lung)}  \right] $.\\ \emph{Number of data-views}: $\textbf{3}$: (a) genomics ($p_3 = \mathit{10299}$),  (b) epigenomics ($p_2 = \mathit{22503}$) and (c) transcriptomics ($p_3 = \mathit{302}$).\\ \emph{Number of samples}: $\textbf{253}$: $\mathit{65}$ patients with breast cancer, $\mathit{82}$ with kidney cancer and $\mathit{106}$ with lung cancer.


\textbf{Reuters \footnote{\href{https://github.com/lzu-cvpr/multiview-learning/blob/master/multiview_DataSets.md}{https://github.com/lzu-cvpr/multiview-learning/blob/master/multiview\_DataSets.md}} \cite{reutersData}:} Text documents.\\ \emph{Number of classes}: $\textbf{6} \left[ \text{E21, CCAT, M11, GCAT, C15, ECAT}  \right] $.\\ \emph{Number of data-views}: $\textbf{5}$: Words from the original documents (English) and from four translations ((a) Italian, (b) French, (c) German and (d) Spanish). All five data-views contain $2000$ features (words).\\ \emph{Number of samples}: $\textbf{1200}: \left[ \textit{200} \text{ documents per class }\right]$. . 	


\textbf{Noisy Data-view Synthetic data (NDS)\cite{rodosthenous2021}:} Synthetic dataset.\\ \emph{Number of classes}: $\textbf{3} \left[\textbf{A}, \textbf{B}, \textbf{C} \right] $.\\ \emph{Number of data-views}: $\textbf{4}$: $p_1 = 100$, $p_2 = 100$, $p_3 = 100$, $p_4 = 1000$ \\ \emph{Number of samples}: $\textbf{300}: \left[ \textit{100} \text{ samples per class }\right]$. 

The synthetic dataset (NDS) aims to justify the use of the algorithm and its ability to capture the true underlying classes of the samples, even when they are not well-represented in each data-view.  The data follow the noisy data-view scenario described by Rodosthenous et al. (2021) \cite{rodosthenous2021} and were generated as follows.  Each sample follows a normal distribution with mean $\mu$ and standard deviation $\sigma = 1$. To distinguish the classes, different $\mu$ values were used for each data-view. Further, noise ($\epsilon \sim \mathcal{N} (\mu_{\epsilon}, \sigma_{\epsilon})$) was added to increase randomness within the data-views. Lastly, polynomial functions were applied on the samples to express non-linearity and ensure that linear dimensionality reduction methods (\textit{e.g.} PCA) would not succeed in identifying the classes. 

In NDS, the first data-view separates only cluster \textbf{A} from the others, the second view separates only cluster \textbf{B}  and the third view separates only cluster \textbf{C}. The first three data-views have $p_v=100$ features. The last data-view represents a noisy data-view (all data points lie in one cluster) with $p_v=1000$ features to intensify the noise.  The data structure in NDS highlights the importance of multi-view analysis, since each data-view describes a distinct clustering, none of which describes accurately the synthetic truth. An effective multi-view algorithm would distinguish the three clusters while it avoids the noise of the $4^{th}$ data-view.



\subsection{Performance evaluation} \label{evalMeasures}

The classification performance of each method, was evaluated by three common measures: (A) Accuracy, (B) Precision, and (C) Recall. Let $T \in \mathbb{R}^{N \times k}$ denote the true classes of a dataset with $N$ samples. For each $1 \leq i \leq N$, 
$T_i \in \mathcal{T} = \{0,1\}^k$, with $1$ referring to its true class. Similarly, let $P \in \mathbb{R}^{N \times k}$ denote the predicted classes. 	\\

\textbf{Accuracy:} The proportion of correctly predicted labels to the total number of labels (predicted and actual).
\begin{align}
	\text{Accuracy} = \frac{1}{N} \sum_{i=1}^{N} \frac{|T_i \cap P_i|}{|T_i \cup P_i|} 
\end{align}

\textbf{Precision:} The proportion of correctly predicted labels to the total number of actual labels.
\begin{align}
	\text{Precision} = \frac{1}{N} \sum_{i=1}^{N} \frac{|T_i \cap P_i|}{|T_i|} 
\end{align}

\textbf{Recall:} The proportion of correctly predicted labels to the total number of predicted labels.
\begin{align}
	\text{Recall} = \frac{1}{N} \sum_{i=1}^{N} \frac{|T_i \cap P_i|}{|P_i|} 
\end{align}

All three evaluation measures lie in the range $[0,1]$, with $0$ referring to a complete misclassification of the unlabelled samples, while $1$ refers to a classification that is perfectly aligned with the ground truth.

\section{Results} \label{results}

In this section, the performance results of S-multi-SNE are presented. Section \ref{results_S_multiSNE} compares several classification algorithms to assess whether the choice of a classifier influences the performance of the algorithm. In Section \ref{resuts_semi_supervised}, S-multi-SNE is compared against LMSSC and St-SNE to explore the performance of our proposal against recent related semi-supervised algorithms. Both LMSSC and St-SNE require parameter tuning. The results presented in this section are obtained with optimal tuning parameters, which were selected according to their respective publications. Section \ref{imbalanced_sss} investigates two specific scenarios, commonly observed in real scenarios: (i) imbalanced data, and (ii) datasets with small sample size. 

\subsection{Classifier selection} \label{results_S_multiSNE}

\begin{table*}[tb]
	\caption{\textbf{Semi-supervised classification.} The mean (and standard deviation) accuracy on bootstrap resamples with different training rates from handwritten digits, Reuters, caltech7 and cancer types data. \textbf{Bold} highlights the method with the best performance on each training rate within each dataset. \textit{BSV St-SNE} refers to the best single-view performance by St-SNE, while \textit{Concat. St-SNE} implements St-SNE on the concatenated features of all data-views. }
	\label{table:comparison_training}
	\centering
	\scalebox{0.80}{
		\begin{tabular}{crcccc}
			\multirow{2}{*}{Dataset} & \multirow{2}{*}{Algorithm} & Training rate & Training rate & Training rate & Training rate \\
			& & 10\% & 20\% & 50\% & 80\% \\				
			\specialrule{0.2em}{0.2em}{0.2em}
			\multirow{2}{*}{Handwritten} &  $\text{S-multi-SNE}^{*}$ & 0.952 \small{(0.016)} & 0.966 \small{(0.008)} & 0.983 \small{(0.004)} & \textbf{ 0.991  \small{(0.004)}} \\
			& LMSSC &  \textbf{0.978 \small{(0.002)}} & \textbf{0.983 \small{(0.003)}} &  \textbf{0.989  \small{(0.002)}} & \textbf{0.991 \small{(0.004)}} \\
			& $\text{BSV St-SNE}^{*}$ & 0.682 \small{(0.022)} & 0.745 \small{(0.027)} & 0.803 \small{(0.025)} & 0.866 \small{(0.014)} \\
			& $\text{Concat. St-SNE}^{*}$ & 0.713 \small{(0.026)} & 0.812 \small{(0.019)} & 0.855 \small{(0.022)} & 0.919 \small{(0.020)}\\
			\addlinespace
			\cdashline{1-6}			
			\addlinespace
			\multirow{2}{*}{Reuters} &  $\text{S-multi-SNE}^{*}$ & 0.554 \small{(0.019)} & 0.632 \small{(0.019)} & 0.767 \small{(0.017)} & \textbf{0.906  \small{(0.018)}} \\
			& LMSSC &  \textbf{0.589 \small{(0.025)} }& \textbf{0.654 \small{(0.022)}} & \textbf{0.857  \small{(0.017)}} & 0.899 \small{(0.012)} \\
			& $\text{BSV St-SNE}^{*}$ & 0.173 \small{(0.058)} & 0.281 \small{(0.030)} & 0.498 \small{(0.028)} & 0.657 \small{(0.051)} \\
			& $\text{Concat. St-SNE}^{*}$ & 0.175 \small{(0.54)} & 0.298 \small{(0.041)} & 0.526 \small{(0.032)} & 0.753 \small{(0.048)} \\			
			\addlinespace
			\cdashline{1-6}			
			\addlinespace
			\multirow{2}{*}{Caltech7} &  $\text{S-multi-SNE}^{*}$ &  \textbf{ 0.924 \small{(0.010)}} & \textbf{0.935 \small{(0.006)}} & \textbf{ 0.961  \small{(0.007)}} & \textbf{0.981  \small{(0.008)}} \\
			& LMSSC & 0.829 \small{(0.040)} & 0.852 \small{(0.019)} & 0.878 \small{(0.011)} & 0.889 \small{(0.011)} \\
			& $\text{BSV St-SNE}^{*}$ & 0.768 \small{(0.014)} & 0.790 \small{(0.19)} & 0.823 \small{(0.022)} & 0.878 \small{(0.20)} \\
			& $\text{Concat. St-SNE}^{*}$ & 0.798 \small{(0.016)} & 0.817 \small{(0.14)} & 0.845 \small{(0.015)} & 0.890 \small{(0.16)} \\			
			\addlinespace
			\cdashline{1-6}
			\addlinespace
			\multirow{2}{*}{Caltech7-balanced} &  $\text{S-multi-SNE}^{*}$ & \textbf{0.928  \small{(0.012)}} &  \textbf{0.950  \small{(0.007)}} & \textbf{0.977  \small{(0.006)}} & \textbf{ 0.991 \small{(0.006)}} \\
			& LMSSC & 0.629 \small{(0.025)} & 0.719 \small{(0.032)} & 0.804 \small{(0.027)} & 0.843 \small{(0.047)} \\
			& $\text{BSV St-SNE}^{*}$ & 0.492 \small{(0.005)} & 0.567 \small{(0.006)} & 0.722 \small{(0.006)} & 0.868 \small{(0.006)} \\
			& $\text{Concat. St-SNE}^{*}$ & 0.515 \small{(0.010)} & 0.681 \small{(0.009)} & 0.819 \small{(0.011)} & 0.910 \small{(0.007)} \\			
			\addlinespace
			\cdashline{1-6}
			\addlinespace
			\multirow{2}{*}{CancerTypes} &  $\text{S-multi-SNE}^{*}$ & 0.661 \small{(0.042)} & 0.769 \small{(0.030)} & 0.914 \small{(0.024)} & \textbf{0.977 \small{(0.017)}} \\
			& LMSSC &  \textbf{0.783  \small{(0.036)}} & \textbf{0.883  \small{(0.038)}} & \textbf{0.957 \small{(0.015)}} & 0.973  \small{(0.012)}\\
			& $\text{BSV St-SNE}^{*}$ & 0.318 \small{(0.044)} & 0.568 \small{(0.021)} & 0.688 \small{(0.034)} & 0.792 \small{(0.028)} \\
			& $\text{Concat. St-SNE}^{*}$ & 0.290 \small{(0.052)} & 0.442 \small{(0.054)} & 0.656 \small{(0.055)} & 0.774 \small{(0.048)} \\			
		\end{tabular}
	}
\end{table*}

The different classification algorithms mentioned in Section \ref{classifiers} are assessed on their predictive performances to evaluate the impact of the classifier on S-multi-SNE.

Following the process described in Section \ref{classifiers}, all six classifiers performed well on every dataset, with SVM, RF and KNN being the most consistent (Table \ref{table:classification_allData} depicts their evaluation performance with 50\% training samples). In particular, KNN performed equally well, or outperformed the other classifiers on all datasets. 

\begin{figure*}[tb]
	\centering
	\resizebox{1.0\textwidth}{!}{\includegraphics[width=\textwidth]{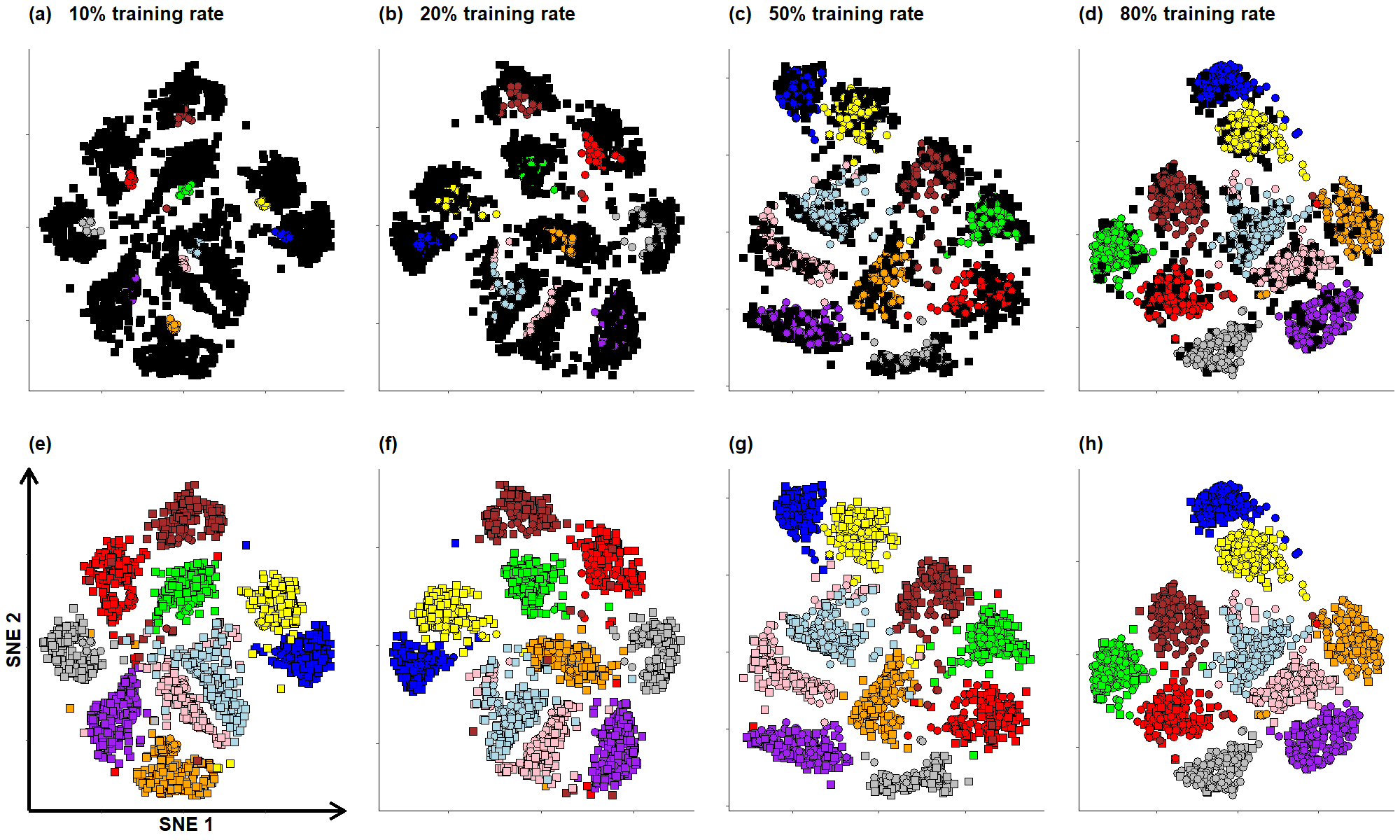}}
	\caption{\textbf{Data visualisation of handwritten digits with different training rates.} \textbf{(a-d)} Unlabelled samples are presented with black squares. \textbf{(e-h)} True labels. The training rates are as follows: \textbf{(a)},\textbf{(e)} 10\%, \textbf{(b)},\textbf{(f)} 20\%, \textbf{(c)},\textbf{(g)} 50\%, and \textbf{(d)},\textbf{(h)} 80\%.}
	\label{fig:reuters_handwritten_training_plots}
\end{figure*}

On the other hand, LDA had the most inconsistent performance across the datasets. This observation is particularly noted on Reuters and caltech7, for which classification is a more challenging task than for the other datasets. Further, DT and NN had the highest variability in performance on all three measures. 

\begin{figure*}[tb]
		\centering
	\resizebox{1.0\textwidth}{!}{\includegraphics[width=\textwidth]{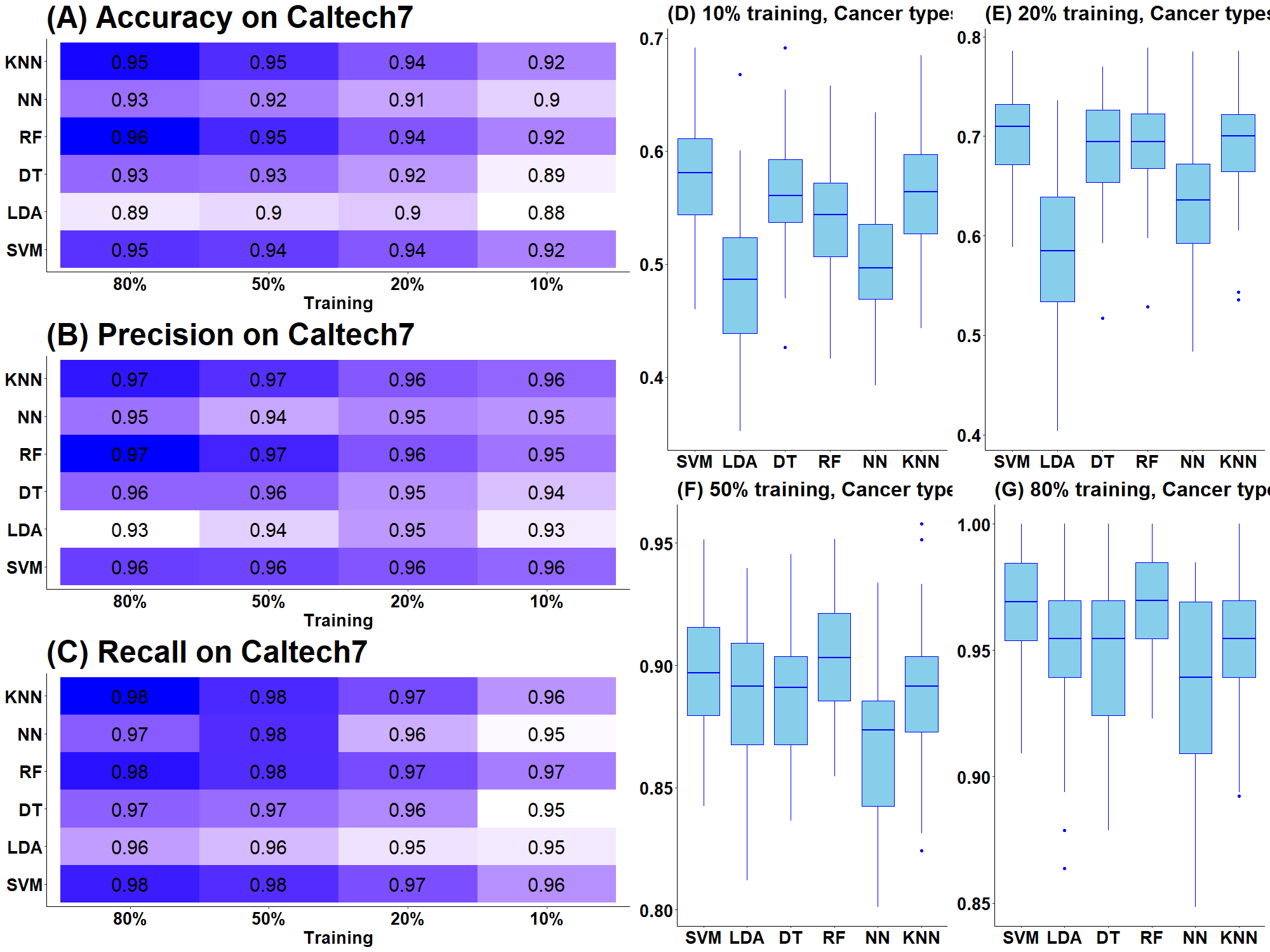}}
	\caption{\textbf{Classifiers performance on caltech7 and cancer types.} (A-C) Heatmaps of accuracy, precision and recall, respectively, on caltech7 with different training samples. (D-G) Box-plots of accuracy on cancer types with 10\%, 20\%, 50\%, 80\% training rates, respectively. On each plot, the performances of all six classifiers are depicted.}  
	\label{fig:boxplot_heatmaps_cal7_cancerT_ggPlot3}
\end{figure*}

Additionally, the performance of the algorithms improve as the number of training samples increases, and all classifiers tend to have similar performances (Figure \ref{fig:boxplot_heatmaps_cal7_cancerT_ggPlot3}). This observation does not come as a surprise, since it is reasonable to expect better classification performance with a larger training set.

Overall, we found KNN to have the most consistently good performance. For that reason, the classification task for the experiments that follow was performed by KNN. Based on the foundation of KNN, \emph{i.e.} finding neighbourhoods in the sample space, and since the embeddings are two-dimensional, there is a direct relationship between its quantitative (classification) and qualitative (visualisation) performances. A good quantitative performance from KNN suggests a good two-dimensional projection of the data, since nearest neighbours belong to the same class.  	

\subsection{Semi-supervised classification} \label{resuts_semi_supervised}

\begin{table}[th]
	\caption{\textbf{Classifiers performance.} The mean (and standard deviation) accuracy, precision and recall of SVM, LDA, DT, RF, NN and KNN with $50\%$ training bootstrap resamples from the handwritten digits, caltech7, cancer types, Reuters and NDS datasets. For each evaluation measure, the classifier with the best performance on each dataset is highlighted with \textbf{bold}.}
	\label{table:classification_allData}
	\centering
	\scalebox{0.85}{
		\renewcommand{\arraystretch}{1.0} 
		\begin{tabular}{c@{\hspace{0.5cm}}r@{\hspace{0.25cm}}c@{\hspace{0.25cm}}c@{\hspace{0.25cm}}c@{\hspace{0.25cm}}c@{\hspace{0.25cm}}c}
			& \multirow{2}{*}{Classifier} & Handwritten  & \multirow{2}{*}{Caltech7} & Cancer & \multirow{2}{*}{Reuters} & \multirow{2}{*}{NDS}  \\
			&  & digits  &  & types &  &   \\
			\specialrule{0.2em}{0.2em}{0.2em}
			\multirow{4}{*}{Accuracy} & SVM & 0.97 \small{(0.005)}  & 0.94 \small{(0.009)} & 0.90 \small{(0.024)} & 0.74 \small{(0.029)} & 0.95 \small{(0.005)} \\
			& LDA & 0.97 \small{(0.008)}  & 0.90 \small{(0.018)} & 0.89 \small{(0.027)} & 0.54 \small{(0.093)} & 0.92 \small{(0.005)} \\
			& DT & 0.95 \small{(0.011)}  & 0.93 \small{(0.011)} & 0.87 \small{(0.029)} & 0.72 \small{(0.025)} & 0.91 \small{(0.013)} \\
			& RF & 0.97 \small{(0.006)}  & \textbf{0.95 \small{(0.007)}} & \textbf{0.91 \small{(0.024)}} & 0.73 \small{(0.021)} & 0.92 \small{(0.007)} \\
			& NN & 0.73 \small{(0.060)}  & 0.92 \small{(0.009)} & 0.89 \small{(0.025)} & 0.71 \small{(0.041)} & 0.92 \small{(0.006)} \\
			& \textbf{KNN} & \textbf{0.98 \small{(0.005)}}  & \textbf{0.95 \small{(0.009)}} & 0.90 \small{(0.023)} & \textbf{0.75 \small{(0.021)}} & \textbf{0.96 \small{(0.004)}} \\
			\addlinespace			
			\cdashline{2-7}			
			\addlinespace		
			\multirow{4}{*}{Precision} & SVM & 0.98 \small{(0.004)}  & 0.96 \small{(0.010)} & \textbf{0.96 \small{(0.019)}} & \textbf{0.86 \small{(0.031)}} & \textbf{0.98 \small{(0.004)}} \\
			& LDA & 0.98 \small{(0.005)}  & 0.94 \small{(0.020)} & 0.95 \small{(0.021)} & 0.70 \small{(0.095)} & 0.94 \small{(0.004)} \\
			& DT & 0.97 \small{(0.007)}  & 0.96 \small{(0.011)} & 0.93 \small{(0.028)} & 0.84 \small{(0.043)} & 0.94 \small{(0.009)} \\
			& RF & 0.98 \small{(0.005)}  & 0.97 \small{(0.007)} & 0.95 \small{(0.020)} & 0.85 \small{(0.020)} & 0.96 \small{(0.005)} \\
			& NN & 0.83 \small{(0.061)}  & 0.94 \small{(0.014)} & 0.95 \small{(0.020)} & 0.84 \small{(0.046)} & 0.94 \small{(0.005)} \\
			& \textbf{KNN} & \textbf{0.99 \small{(0.004)}}  & \textbf{0.97 \small{(0.008)}} & \textbf{0.96 \small{(0.017)}} & \textbf{0.86 \small{(0.024)}} & \textbf{0.98 \small{(0.004)}} \\
			\addlinespace
			\cdashline{2-7}		
			\addlinespace					
			\multirow{4}{*}{Recall} & SVM & \textbf{0.98 \small{(0.005)}}  & \textbf{0.98 \small{(0.007)}} & \textbf{0.94 \small{(0.021)}} & 0.84 \small{(0.030)} & \textbf{0.98 \small{(0.003)}} \\
			& LDA & \textbf{0.98 \small{(0.007)}}  & 0.96 \small{(0.011)} & 0.93 \small{(0.025)} & 0.71 \small{(0.098)} & 0.97 \small{(0.003)} \\
			& DT & 0.97 \small{(0.010)}  & 0.97 \small{(0.011)} & 0.93 \small{(0.028)} & 0.83 \small{(0.035)} & 0.97 \small{(0.010)} \\
			& RF & \textbf{0.98 \small{(0.005)}}  & \textbf{0.98 \small{(0.011)}} & 0.93 \small{(0.021)} & \textbf{0.85 \small{(0.021)}} & 0.97 \small{(0.005)} \\
			& NN & 0.87 \small{(0.092)}  & \textbf{0.98 \small{(0.011)}} & 0.93 \small{(0.023)} & 0.82 \small{(0.043)} & 0.97 \small{(0.004)} \\
			& \textbf{KNN} & \textbf{0.98 \small{(0.004)}}  & \textbf{0.98 \small{(0.011)}} & \textbf{0.94 \small{(0.019)}} & \textbf{0.85 \small{(0.022)}} & \textbf{0.98 \small{(0.002)}}  
		\end{tabular}
	}
\end{table}




A comparative study between S-multi-SNE, LMSSC and St-SNE was conducted	to assess the performance of S-multi-SNE against similar state-of-the-art algorithms. The aim of this section is two-fold. The comparison with LMSSC demonstrates the performance of S-multi-SNE against a state-of-the-art multi-view semi-supervised classification method, while the comparison with St-SNE highlights the benefits of incorporating multiple data-views in contrast to single-view classification and visualisation.

S-multi-SNE outperformed St-SNE in any combination of datasets and training rates (Table \ref{table:comparison_training}). St-SNE is not a semi-supervised approach and a good performance on low training rates was not expected. On 80\% training rate, the performance of St-SNE did not match the ones from LMSSC or S-multi-SNE which can be explained by the lack of multiple data-views. Concatenating the features of all data-views before implementing St-SNE does not improve its performance. Fu et al. (2008) \cite{fu2008} argue that this is because the information conveyed by different features is not equally represented, since the data-views are described by different data distributions and variation patterns.

\begin{figure*}[tbh]
	\centering
	\resizebox{1.0\textwidth}{!}{\includegraphics[width=\textwidth]{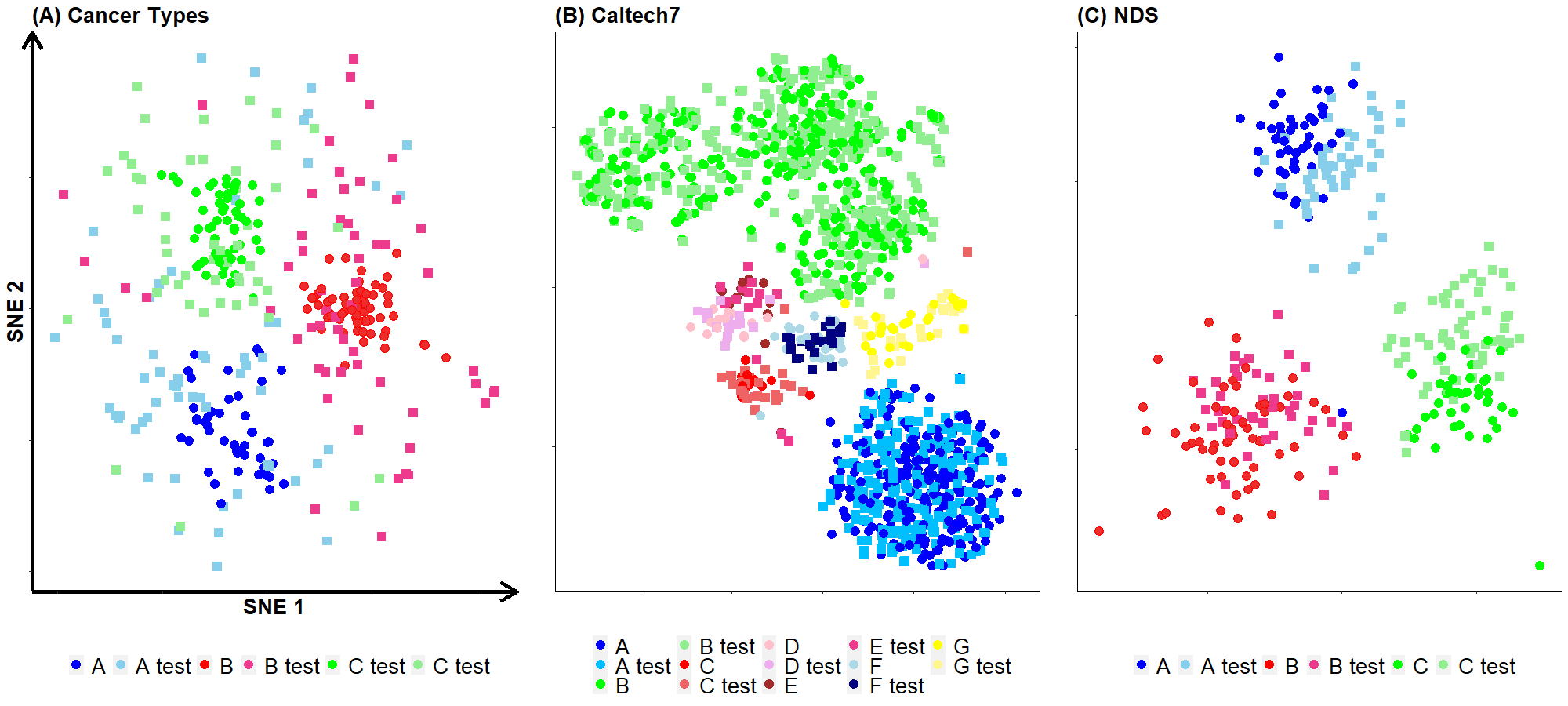}}
	\caption{\textbf{Data visualisation with 50\% training samples.} (A) Cancer types, (B) caltech7 and (C) NDS projections of S-multi-SNE. The training samples are presented with circles and test samples with squares.}
	\label{fig:multiSNE_visualisation_nds_caltech7_cancerT}
\end{figure*}

As expected, the accuracy of all methods increases when more training samples are available (Table \ref{table:comparison_training}). On handwritten digits, Reuters and cancer types, and on low training rates ($10\%,20\%,50\%$) the performance of S-multi-SNE was slightly lower, but still comparable to LMSSC. However, with $80\%$ training rate, S-multi-SNE surpasses the performance of LMSSC. 
The performance of the methods is influenced by the quality of the data, especially on low training rates (Figures \ref{fig:reuters_handwritten_training_plots} and \ref{fig:multiSNE_visualisation_nds_caltech7_cancerT} ). When training is performed on just $10\%$ of the samples, S-multi-SNE projects the unlabelled samples in Reuters mostly as noise instead of signal (Figure 2 in supplementary material), whereas in handwritten digits, the distinction between classes is successfully achieved (Table \ref{table:comparison_training} and Figure \ref{fig:reuters_handwritten_training_plots}).


The analysis on cancer types agrees with the conclusions made on handwritten digits and Reuters. On the other hand, S-multi-SNE on caltech7 outperforms LMSSC on all training rates (Table \ref{table:comparison_training}). The split between training and test was performed proportionally to the class sizes. This means that with $10\%$ training, we would only have $4-6$ labelled samples for five out of seven classes. S-multi-SNE overcomes this challenge with a better accuracy than LMSSC. This observation could suggest that S-multi-SNE is more robust than LMSSC on imbalanced datasets. 

To investigate this claim further, a balanced subset of caltech7 was created, by reducing the sample size of classes \textit{A} and \textit{B} to $50$ samples. Even though this under-sampling process defeats the obstacle of imbalanced samples, a new challenge arises; overall, a small sample size per class is observed. This new challenge reduced the accuracy of LMSSC, while the performance of S-multi-SNE was unaffected (agrees with the performance of multi-SNE on a similar experiment \cite{rodosthenous2021}). This observation could suggest that S-multi-SNE can be used more effectively than LMSSC when the dataset contains a small sample size per class.


\subsection{Imbalanced and small sample size} \label{imbalanced_sss} 

The comparison study between S-multi-SNE and LMSSC in the previous section and specifically on caltech7 dataset, suggests that the former algorithm is more effective than the latter, when the dataset has imbalanced labels or each class has a small number of samples. To explore these two hypotheses further, two subsets of NDS were taken: (A) $NDS_{im}$: Represents an imbalanced dataset, with $150$ samples; all $100$ samples from class A were part of the subset, while $20$ and $30$ samples were randomly selected from classes B and C, respectively. (B) $NDS_{sss}$: Represents a small sample size dataset, with $30$ samples; $10$ samples per class were randomly selected.

S-multi-SNE outperformed LMSSC on the synthetic dataset and its subsets (Table \ref{table:imbalanced_sss}). This performance can be explained by the noise of the $4^{th}$ data-view, which had a bigger influence on LMSSC than on S-multi-SNE. Both algorithms had a similar and comparable classification scores on both $NDS_{im}$ and $NDS_{sss}$. On all scenarios, S-multi-SNE was slightly more accurate than LMSSC, except on $NDS_{sss}$ with 50\% training rate.

Throughout all experiments, S-multi-SNE classified the test samples more accurately, when 80\% of the samples were in training, but for the remaining training rates, the performances of S-multi-SNE and LMSSC were interchangeable. Although, none of the algorithms showed an evident advantage over the other in the classification task, S-multi-SNE has the benefit of providing an auxiliary comprehensible projection of all samples. This feature may be desirable by researchers who want to explore visually the classes within their data, in addition to the corresponding classification predictions.

\begin{table*}[bh]
	\caption{\textbf{Imbalanced and small sample size classification.} The mean (and standard deviation) accuracy on bootstrap resamples with different training rates on NDS, $NDS_{im}$ and $NDS_{sss}$. \textbf{Bold} highlights the algorithm with the best performance on each training rate within each dataset. }
	\label{table:imbalanced_sss}
	\centering
	\scalebox{1.0}{
		\begin{tabular}{crcccc}
			\multirow{2}{*}{Dataset} & \multirow{2}{*}{Algorithm} & Training rate & Training rate & Training rate & Training rate \\
			& & 10\% & 20\% & 50\% & 80\% \\				
			\specialrule{0.2em}{0.2em}{0.2em}
			\multirow{2}{*}{NDS} &  S-multi-SNE & \textbf{0.884 \small{(0.03)}} & \textbf{0.903 \small{(0.02)}} & \textbf{0.960 \small{(0.09)}} &\textbf{ 0.984 \small{(0.03)}} \\
			& LMSSC &  0.714 \small{(0.05)} &  0.733 \small{(0.04)} & 0.887  \small{(0.07)} &  0.923 \small{(0.05)} \\
			\addlinespace
			\cdashline{1-6}			
			\addlinespace
			\multirow{2}{*}{$NDS_{im}$} & S-multi-SNE & \textbf{0.686 \small{(0.03)}} &  \textbf{0.738 \small{(0.04)}} & \textbf{0.775  \small{(0.06)}} & \textbf{0.833 \small{(0.04)}} \\				
			& LMSSC  &  0.669 \small{(0.02)} & 0.670 \small{(0.03)} & 0.767 \small{(0.05)} & 0.800 \small{(0.09)}\\
			\addlinespace
			\cdashline{1-6}			
			\addlinespace
			\multirow{2}{*}{$NDS_{sss}$} &  S-multi-SNE & \textbf{0.354 \small{(0.03)}} & \textbf{0.455 \small{(0.07)}} & 0.589 \small{(0.08)} &  \textbf{0.797 \small{(0.13)}} \\
			& LMSSC &  0.345 \small{(0.06)} & 0.412 \small{(0.08)} &  \textbf{0.673 \small{(0.05)}} &  0.705 \small{(0.09)} \\								
		\end{tabular}
	}
\end{table*}

\section{Discussion} \label{discussion}

In this work, we propose S-multi-SNE, a semi-supervised learning algorithm for data visualisation and classification. S-multi-SNE produces low-dimensional projections which are used as input features in a classification algorithm to classify the test samples. 	Although, we found that the classifiers do not have a big effect on the performance of S-multi-SNE, KNN produced the most consistently good predictions out of all six standard classification algorithms tested in this manuscript. Compared against a state-of-the-art multi-view semi-supervised classification approach, S-multi-SNE performed equally well. Specifically,  it outperformed LMSSC on caltech7, synthetic dataset NDS, and their subsets, which cover two scenarios: (a) Balanced against imbalanced samples, and (b) small number of samples per class. In addition to its strong classification performance, S-multi-SNE has the desirable feature of producing a comprehensible projection that splits all samples (training and test) to their corresponding classes. On all datasets, S-multi-SNE outperformed a recently proposed supervised variation of t-SNE. This comparison emphasizes on the benefits and importance of performing multi-view analysis over single-view, when available. 



Although it was not tested explicitly in this study, S-multi-SNE can be applied on single-view data as well. In this scenario, two data-views would be considered: the first (\emph{i.e.} $X^{(1)}$) would represent the single-view dataset and the second would contain the labelling information (\emph{i.e.} $X^{(2)} = X^{(l)}$).

In some real datasets, it is possible to have samples with missing information in one or more data-views. This can be a result of technical, human or other errors. For example, in a study on genomics, transcriptomics and epigenomics, experimentalists may be unable to get the epigenomics measurements from several patients, but they can have transcriptomics and genomics data. Note that the entire information of a sample would be missing, and not just some values from selected features that got lost. In such situations, a sample with missing information is often entirely excluded from a multi-view analysis, since such analyses require the same number of samples from all data-views. The algorithm of S-multi-SNE can be generalised to allow the analysis of data-views with missing samples. We refer to this generalisation as \textit{G-multi-SNE} and its cost function is given by: 

\begin{align}
	\small{
		C_{G-multiSNE} = \sum_m \sum_{i} \sum_{j}  \mathfrak{I}^{(m)}_{ij} w^{(m)} p^{(m)}_{ij} \log \frac{p^{(m)}_{ij}}{q_{ij}} ,
	}
\end{align}
where  $\mathfrak{I}^{(m)}_{ij} = \mathds{1} \{ \{D^{(m)}_{i} = 0 \} \land \{D^{(m)}_{j} = 0 \} \} \text{ and }D^{(m)}_{i} = \mathds{1} \{ \textbf{x}_i^{(m)} \text{ is missing} \} \in \mathbb{R}^{N}, \quad \forall m=1, \cdots M$.

G-multi-SNE, has the potential of increasing the overall sample size, by including, instead of ignoring, samples with missing information. By implementing this approach, researchers could classify the unlabelled samples and at the same time visualise them along with labelled samples and samples with missing information (a visualisation example of G-multi-SNE applied on NDS is displayed in supplementary material).  


In this study, we have shown that S-multi-SNE can perform comparably well with a state-of-the-art semi-supervised multi-view classification method, while producing a comprehensive visualisation on two dimensions. 

\section*{Reproducibility}
The public multi-view datasets used in this manuscript can be found by following the links provided in the main body of the paper. We refer the readers to follow the code and functions provided in the link below to reproduce the findings of this paper: \url{https://github.com/theorod93/S_multi_SNE}.\\
The \texttt{R} package \texttt{multiSNE} contains the code and functions required to run both multi-SNE and S-multi-SNE. It can be installed through \texttt{GitHub} (and \texttt{devtools}) from the repository found in: \url{https://github.com/theorod93/multiSNE}.

	\clearpage
	\vskip 0.2in
	\bibliographystyle{unsrtnat}

\end{document}